\documentclass[conference]{IEEEtran}
\IEEEoverridecommandlockouts

\usepackage{cite}
\usepackage{amsmath,amssymb,amsfonts}
\usepackage{algorithmic}
\usepackage{graphicx}
\usepackage{textcomp}
\usepackage{xcolor}
\def\BibTeX{{\rm B\kern-.05em{\sc i\kern-.025em b}\kern-.08em
    T\kern-.1667em\lower.7ex\hbox{E}\kern-.125emX}}
\begin{document}

\title{Quantitative evaluation of brain-inspired vision sensors in high-speed robotic perception
}




\author{\IEEEauthorblockN{ Taoyi Wang$^{*}$, Lijian Wang$^{*}$, Yihan Lin, Mingtao Ou, Yuguo Chen, Xinglong Ji, Rong Zhao}
\IEEEauthorblockA{\textit{Center for Brain-Inspired Computing Research (CBICR)} \\
\textit{Department of Precision Instrument, Tsinghua University}\\
Beijing, China \\
\{wangty23, wlj24, linyh20, cyg22\}@mails.tsinghua.edu.cn, \{mingtao-ou, xinglongji, r\_zhao\}@mail.tsinghua.edu.cn }
\thanks{*These authors equally contribute to this work.}
\and


}

\maketitle

\begin{abstract}
Perception systems in robotics encounter significant challenges in high-speed and dynamic conditions when relying on traditional cameras, where motion blur can compromise spatial feature integrity and task performance. Brain-inspired vision sensors (BVS) have recently gained attention as an alternative, offering high temporal resolution with reduced bandwidth and power requirements. Here, we present the first quantitative evaluation framework for two representative classes of BVSs in variable-speed robotic sensing, including event-based vision sensors (EVS) that detect asynchronous temporal contrasts, and the primitive-based sensor Tianmouc that employs a complementary mechanism to encode both spatiotemporal changes and intensity. A unified testing protocol is established, including cross-sensor calibrations, standardized testing platforms, and quality metrics to address differences in data modality. From an imaging standpoint, we evaluate the effects of sensor non-idealities, such as motion-induced distortion, on the capture of structural information. For functional benchmarking, we examine task performance in corner detection and motion estimation under different rotational speeds. Results indicate that EVS performs well in high-speed, sparse scenarios and in modestly fast, complex scenes, but exhibits performance limitations in high-speed, cluttered settings due to pixel-level bandwidth variations and event rate saturation. In comparison, Tianmouc demonstrates consistent performance across sparse and complex scenarios at various speeds, supported by its global, precise, high-speed spatiotemporal gradient samplings. These findings offer valuable insights into the application-dependent suitability of BVS technologies and support further advancement in this area.
\end{abstract}


\section{Introduction}
\begin{figure}[htbp]
    \centering
    \includegraphics[width=1\linewidth]{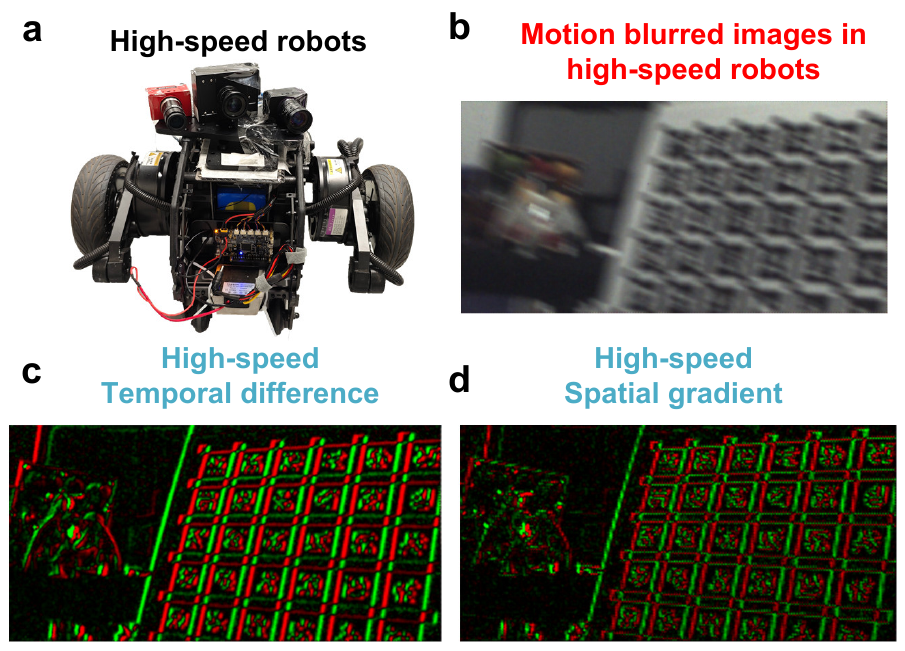} 
    \caption{ (a) A typical high-speed robot platform. (b) Challenges of motion blur in the Tianmouc COP pathway, which shares the same motion blur issues as traditional cameras, when the robot in (a) is turning rapidly. (c) The high-speed temporal difference and (d) spatial difference sampling in the Tianmouc AOP pathway demonstrate significant potential for motion-blur-resistant robotic perception.}
    \label{fig:hschall}
\end{figure}

Recently, brain-inspired vision sensors (BVS) have emerged as promising alternatives for overcoming the  limitations of traditional image sensors under high-speed robotic applications \cite{gallego2020event,yang2024vision}. Among them, Tianmouc \cite{yang2024vision}, event-based sensors (EVS) \cite{ evk4spec, inivation_all_spec}, and RGB hybrid EVS \cite{ brandli2014240, DAVIS346_spec} exhibit strong potential due to their high-speed sampling scheme.  EVS is known for low power and low bandwidth requirement since they only encode temporal changes in visual information as low-latency asynchronous events. RGB hybrid EVS is developed to complement the EVS by providing high-precision sensing of static scenes.
Tianmouc, as an emerging class of BVS based on a primitive-based paradigm, implements two complementary pathways: the cognition-oriented pathway (COP) for accurate colorful intensity sensing and the action-oriented pathway (AOP) for high-precision, fast, and sparse multi-bit temporal difference (TD) and spatial difference (SD) sensing. This complementary dual-pathway architecture offers comprehensive visual information with low bandwidth consumption, and the AOP pathway can provide high-accuracy data without motion blur and temporal aliasing as shown in Fig.\ref{fig:hschall}c-d. Notably, the AOP pathway, particularly through its SD sensing capability, continuously generates robust visual information \cite{he2024microsaccade}. 

Despite the advantages of BVS, selecting the appropriate sensor for specific robotic applications remains challenging due to their novel sensing paradigms, the absence of well-adopted test standards like EMVA1288 \cite{jahne2010emva}, and incompatibility with traditional vision algorithms. While previous studies have examined EVS performance in specific robotic systems and tasks, such as flapping-wing robot \cite{tapia2023comparison} and ultra-high speed video capture \cite{holevsovsky2021experimental}, a comprehensive analysis across different BVS technologies, including the emerging class of primitive-based sensors, has yet to be conducted within the context of robotic applications.

\begin{figure}
    \centering
    \includegraphics[width=0.9\linewidth]{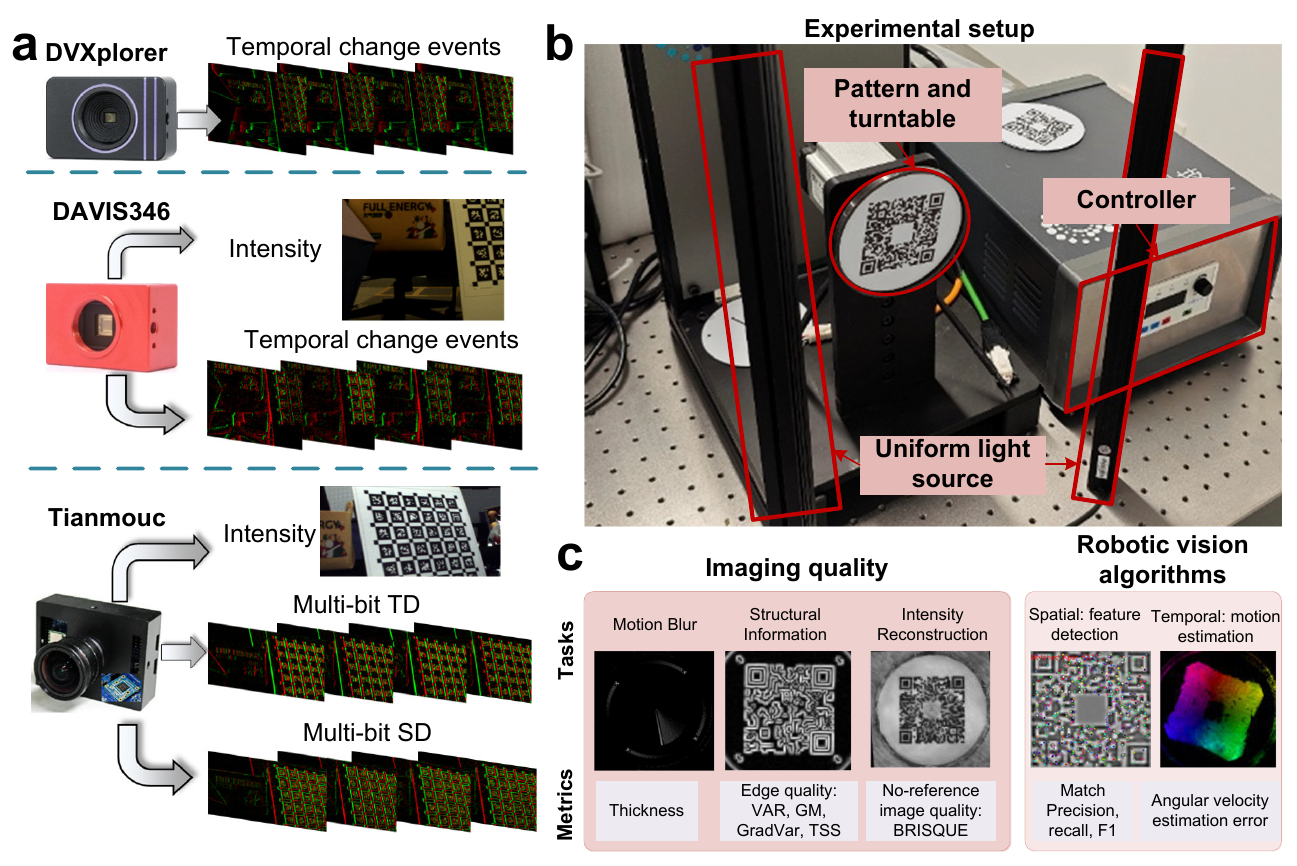} 
    \caption{Overview of our work. (a) Visualization of difference in data modalities of different sensors. (b) The experimental setup. (c) Two key quality factors for evaluation.}
    \label{fig:framework}
\end{figure}

In this paper, we present the first evaluation framework for quantitatively assessing the sensing capabilities of different classes of BVS in speed-dependent robotic perception, with a particular focus on high-speed scenarios  (Fig.\ref{fig:framework}). To ensure fair evaluations, we select DVXplorer as a representative EVS, given its resolution and bandwidth similarity to Tianmouc, and the widely used DAVIS346 as a typical RGB hybrid EVS. Their differing sensing principles are illustrated in Fig.\ref{fig:framework}a. Unlike EVS, which exclusively detects temporal changes, Tianmouc’s AOP simultaneously captures TD and SD, enabling a comprehensive representation of dynamic visual information in both temporal and spatial domains.

To facilitate generalizable and reproducible evaluations, we introduce a unified platform comprising a suite of calibration methods, standardized experimental setups, and well-defined quality metrics for quantitative assessment. The proposed calibration methods compensate for modality differences between sensors by applying motion compensation on event data. The experimental setup (Fig.\ref{fig:framework}b) consists of a turntable with adjustable rotational speeds, a tunable light source, and printed patterns generated by a program as ground truth references. Our evaluation framework focuses on two key quality factors of robotic perception: 1) imaging quality, assessed through the accuracy of edge information capture, which is a fundamental capability for BVS, and 2) functional performance in core vision tasks, including spatial analysis (corner detection) and temporal estimation (motion tracking). Quantitative assessments are conducted across a range of rotation speeds, from 50 to 3000 rounds per minute (rpm), depending on task requirements.

The main contributions of this work are threefold: (1) the establishment of the first standardized, quantitative evaluation framework for assessing BVS performance in robotic perception; (2) the development of a generalizable, quantitative motion-compensated calibration method for robust cross-modality image quality assessment; and (3) a systematic comparison of performance in fundamental vision tasks, specifically, corner detection and motion estimation, across representative BVS technologies. Our results indicate that EVS maintains high data fidelity in sparse, high-speed scenarios and in moderately dynamic, complex environments. However, their performance degrades under cluttered, high-speed conditions due to pixel bandwidth variations and event rate saturation. In contrast, Tianmouc, with its global high-speed and high-precision spatiotemporal gradient sampling, demonstrates robust performance in both sparse and complex scenarios at varying speeds, highlighting its advantages for high-speed robotic perception. These results provide actionable insights into the application-specific suitability of BVS technologies and offer guidance for the future design and deployment of BVS in robotic perception systems.


This paper is structured as follows. Section \ref{sec:relate} reviews the main related works. Section \ref{sec:principle} introduces the sensing principles of different BVS technologies. Section \ref{sec:imging_quality} presents the evaluation methods and results of the imaging quality of BVS. Section \ref{sec:robo_perc} analyzes the performance of BVS  on fundamental robotic perception tasks. Section \ref{sec:discuss} discusses practical implications of BVS across different scenarios, concludes with key findings, and provides future research directions.


\section{Related Works}\label{sec:relate}

\subsection{Analysis of data quality of event cameras}
The application of EVS in robotics and computer vision tasks has surged in recent years due to its low latency, high dynamic range, and low power consumption. However, recent studies have identified various non-idealities that arise in challenging environments, including low-light, high-speed, and highly cluttered environments. For instance, motion blur induced by pixel latency variation under low illuminations has been analyzed in \cite{hu2021v2e}. Additionally, EVS output bus saturation in high-speed, cluttered scenes can lead to severe distortion, reducing the performance of downstream tasks \cite{gehrig2022high}. The impact of blurred event data on perception tasks has been further investigated in \cite{yang2024latency,low2024deblur}. Furthermore, the exclusive reliance of EVS on temporal changes results in information loss when object edges align parallel to the camera motion, limiting their effectiveness in certain scenarios \cite{he2024microsaccade, gallego2020event, gehrig2020eklt}.



\subsection{Benchmarks for event-based systems}
To address these challenges, several benchmarks have been developed for evaluating EVS-based vision algorithms. For low-level vision algorithms, EVREAL \cite{ercan2023evreal} provides a benchmark for event-based video reconstruction, while E-MLB \cite{ding2023mlb} focuses on event stream denoising and introduces new evaluation metrics for advanced denoising algorithms. EventAid \cite{duan2023eventaid} serves as a benchmark for event sensors assisting conventional sensors in image and video enhancement. For high-level vision algorithms, THUE-ACT-50 \cite{gao2023action} offers a large-scale dataset for action recognition, while \cite{wang2024event} introduces a benchmark dataset for object tracking. However, existing benchmarks primarily focus on evaluating event-based vision algorithms, lacking a systematic approach to compare different EVS or various types of BVS.

\subsection{Comparison between EVS and traditional cameras}
Several studies have compared EVS with traditional frame-based cameras. For instance, \cite{tapia2023comparison} evaluates DAVIS346 against conventional cameras in ornithopter applications, suggesting that EVS outperforms traditional sensors in terms of power, size, dynamic range, and motion blur. However, EVS performance remains comparable or even inferior in many vision tasks in \cite{tapia2023comparison}. Another study \cite{holevsovsky2021experimental} compares ultra-high-speed cameras with EVS, concluding that while EVS achieves superior bandwidth efficiency, they remain constrained by pixel latency and output bandwidth, particularly in cluttered and high-speed scenarios. Object tracking comparisons between EVS and traditional cameras in \cite{movementvsDVS} indicate that EVS excels in data efficiency, detection speed, and accuracy. Furthermore, theoretical analysis \cite{censi2015power} suggests that different sensors dominate at varying power-performance trade-offs, emphasizing the need for application-specific sensor selection. Notably, none of these studies systematically compares different EVS or examines the emerging class of BVS.

\section{Principle of various BVS}
\label{sec:principle}
In this section, we introduce the sensing principles of different types of BVS, including visual information sensing principles and core characteristics. Event cameras, as a subset of BVS, primarily sense temporal changes, which have been widely described using the event generation model \cite{gallego2020event}. Given an event $(x_k, y_k, p_k, t_k)$ generated at pixel $(x_k, y_k)$ in a noise-free way ,the event generation model is represented by Eq.\ref{eq:gen_model}:
\begin{equation}
    \label{eq:gen_model}
    \Delta L(x_k, y_k, t_k)=L(x_k, y_k, t_k) - L(x_k, y_k, t_k - \Delta t_k) = p_k C,
\end{equation}
where $p_k$ is the polarity of the event and $C$ is the contrast threshold of the pixel. $L(x, y, t)$ represents the logarithm of the irradiance response $I(x, y, t)$.
Assuming events are triggered by moving edges, their distribution follows:
\begin{equation}
\label{dvs_edge}
     \Delta L(x, y, t) \sim -\nabla L(x, y, t) \cdot v \cdot \Delta t,
\end{equation}
where $v$ is the velocity of objects on the focal plane.

Unlike conventional EVS, the Tianmouc chip \cite{yang2024vision}, with a primitive-based representation, forms two complementary sensing pathways that parse comprehensive visual information with high bandwidth efficiency. The COP captures dense color intensities, while the AOP extracts sparse, multi-bit (up to $\pm$7-bit) temporal difference and spatial difference features at a global sampling rate up to 1515 fps. The TD and SD can be formalized in Eq.\ref{eq:tdsd}, where $I(x,y,t)$ represents the optical intensity at the pixel position $(x,y,t)$. In practical usage, the $SD_{LeftDir}$ and $SD_{RightDir}$ can be easily converted to the spatial gradient in the X and Y axes. 

\begin{flalign} \label{eq:tdsd}
\begin{split}
     &SD_{LeftDir}(x, y, t) = I(x, y, t) - I(x-1, y-1, t)  \\
     &SD_{RightDir}(x, y, t) = I(x, y, t) - I(x+1, y+1, t) \\
     &TD(x, y, t) =  I(x, y, t) - I(x, y, t-1)
\end{split}
 \end{flalign}
 

Despite differences in sampling schemes, both EVS and Tianmouc share similar edge information representations through spatial or temporal change sensing, enabling the development of a unified general methodology for evaluation.

\section{Imaging quality evaluation}
\label{sec:imging_quality}
\subsection{Experimental setups}
\begin{figure*}
    \centering
    \includegraphics[width=1\linewidth]{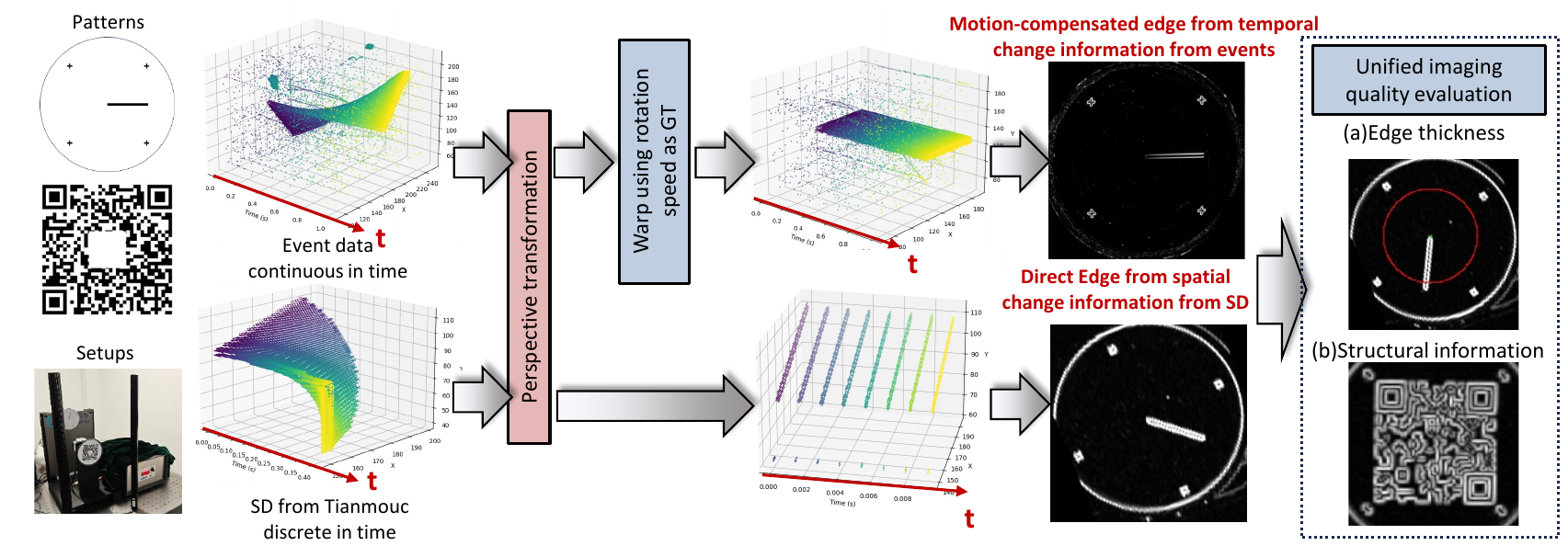} 
    \caption{ Evaluation methods for imaging quality of BVS.}
    \label{fig:img_q_method}
\end{figure*}

Edge representation and reconstruction are critical indicators of imaging quality in BVS, as numerous algorithms rely on abundant edge information for effective processing \cite{iwe2018,eros}. To systematically evaluate camera performance across varying illumination levels, motion speeds, and feature complexity of the targets, we develop a standardized test platform. It integrates adjustable uniform light sources, a mechanical turntable with accurate controllable rotation speeds, and printed patterns that can be customized based on experimental requirements. This setup facilitates a rigorous assessment under varying conditions of illumination, motion speed, and pattern complexities. As shown in Fig.\ref{fig:img_q_method}, each camera is directed toward a turntable capable of rotating at speeds ranging from 0 to 3000 rpm. To assess imaging quality under different light conditions, we maintain a constant low illumination of 100 lux and a high illumination of 2,000 lux.

We adopt DAVIS346 as a representative of the RGB hybrid EVS and DVXplorer as a representative of EVS. To mitigate scan pattern distortion in EVS \cite{inivation_all_spec}, which arises from event rate saturation, we implement an alternative experimental setup for the DVXplorer for a fair comparison. This setup reduces the region of interest (ROI) to match the resolution of other cameras, named DVXPlorer\_ROI, ensuring effective pattern sensing under high-load conditions, such as QR codes. Given the adaptive nature of EVS to illumination changes, we employ their default settings. For Tianmouc, we configure an exposure time of $\geq$1 ms and a frame rate of 757 fps for low illumination and 0.4 ms exposure and 1515 fps for high illumination, respectively. 



\subsection{Data preprocessing}
The inherent asynchrony of EVS poses challenges in edge reconstruction, as simple event accumulation within a fixed time window or event count fails to recover sharp edges accurately. To address this, we adopt a motion-compensated event calibration method based on the rotation speed of the turntable. Specifically, events accumulated over a 15$^{\circ}$ rotation window are transformed via perspective correction, aligning data event-by-event. The transformed events are then warped back using the turntable’s rotation speed as optical flow ground truth, similar to the contrast maximization framework and image of warped events \cite{iwe2018}. This method eliminates the errors associated with direct event accumulation and provides a precise evaluation of edge recovery in EVS. For Tianmouc, which directly captures edge information through SD data, we apply perspective transformation for evaluation without additional processing. This ensures a fair comparison across different BVS modalities. 


\subsection{Motion blur evaluation}
\begin{figure}
    \centering
    \includegraphics[width=0.75\linewidth]{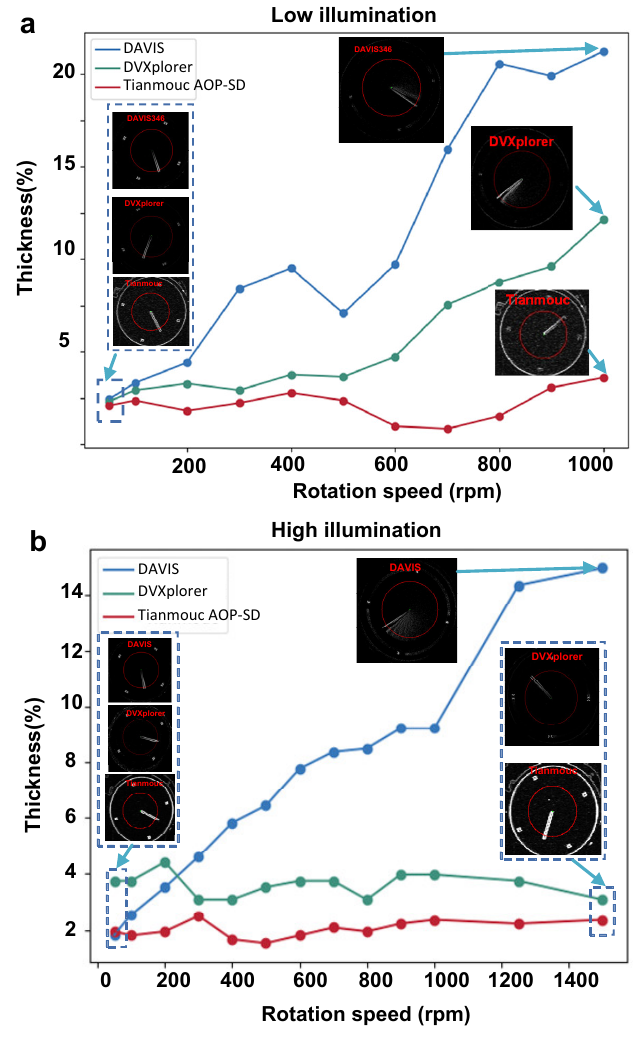} 
    \caption{Evaluation of motion blur using thickness as the indicator under (a) low illumination and (b) high illumination.}
    \label{fig:thick}
\end{figure}

To quantitatively assess edge-sensing accuracy in BVS, we propose two key metrics. The first metric measures the thickness of processed event or SD data by sensing a rotating line to gauge motion blur in relation to varying speeds and illuminations, a method analogous to established motion blur evaluation techniques \cite{tapia2023comparison}. The thickness is determined as the peak width (from maximum intensity to near zero) at a pixel position located at 90\% of the radius from the turntable center in the transformed line edge representation. The computed results are illustrated in Fig.\ref{fig:thick}. Under high illumination, DVXplorer and Tianmouc AOP-SD exhibit strong robustness to speed changes, whereas DAVIS346 demonstrates substantial degradation due to motion-induced artifacts. Under low illumination, both DAVIS346 and DVXplorer exhibit increased edge thickness and motion blur, indicating reduced performance in capturing fine details. In contrast, Tianmouc AOP-SD maintains resilience to speed variations. In low light conditions, higher speeds may cause motion blur in the Tianmouc, DAVIS346, and DVXplorer cameras due to the limited sensitivity of the photosensitive elements in their pixels, which will be analyzed in our future work.


\subsection{Structural information variation}
While the thickness metric effectively evaluates motion blur in BVS under simple and sparse scenarios, it does not fully capture variation in data quality when complex features are present. The data rate significantly influences data quality \cite{yang2016analysis}, leading to structural information degradation, particularly in high-speed and highly cluttered scenarios. This degradation can be quantified by analyzing the accuracy of recovered edge information. To systematically assess structural degradation under extreme conditions, we employ a complex QR code pattern, analogous to AprilTag markers commonly used in robotic vision systems. This pattern simulates real-world robotic perception in dynamic environments, where rapid motion and dense features impose substantial demands on visual processing. We evaluate the structural integrity of edge-like images derived from transformed event and AOP-SD data using four key indicators: gradient magnitude (GM), total sum of squares (TSS), variance (VAR), and variance of the gradient image (GradVar). These metrics provide a quantitative assessment of structural fidelity, similar to gradient-based image quality evaluation methods. TSS is calculated as the sum of the squared calibrated data: 

\begin{equation}
    \mathrm{TSS}:= \sum_{x}I(x)^2.
\end{equation}
The GM can be computed by 
\begin{equation}
    \mathrm{GM}:=\frac{1}{|\Omega|} \cdot \int_{\Omega}\|\nabla I(\boldsymbol{x})\|^{2} \mathrm{~d} \boldsymbol{x},
\end{equation}
where $\Omega$ is the set of all pixel positions in the image space, $I(\boldsymbol{x})$ is the number of events (or intensities) responding to a pixel, and $\mu_{I}$ is the average of the total number of events. The VAR measures the variance of the transformed data, while the GradVar quantifies the variance of the gradient of the transformed data. For our comparison, we employ a relative measurement approach by normalizing the indicators against the values obtained at the lowest speed. This normalization is essential because the absolute indicator values vary significantly across different cameras.

\begin{figure*}
    \centering
    \includegraphics[width=1\linewidth]{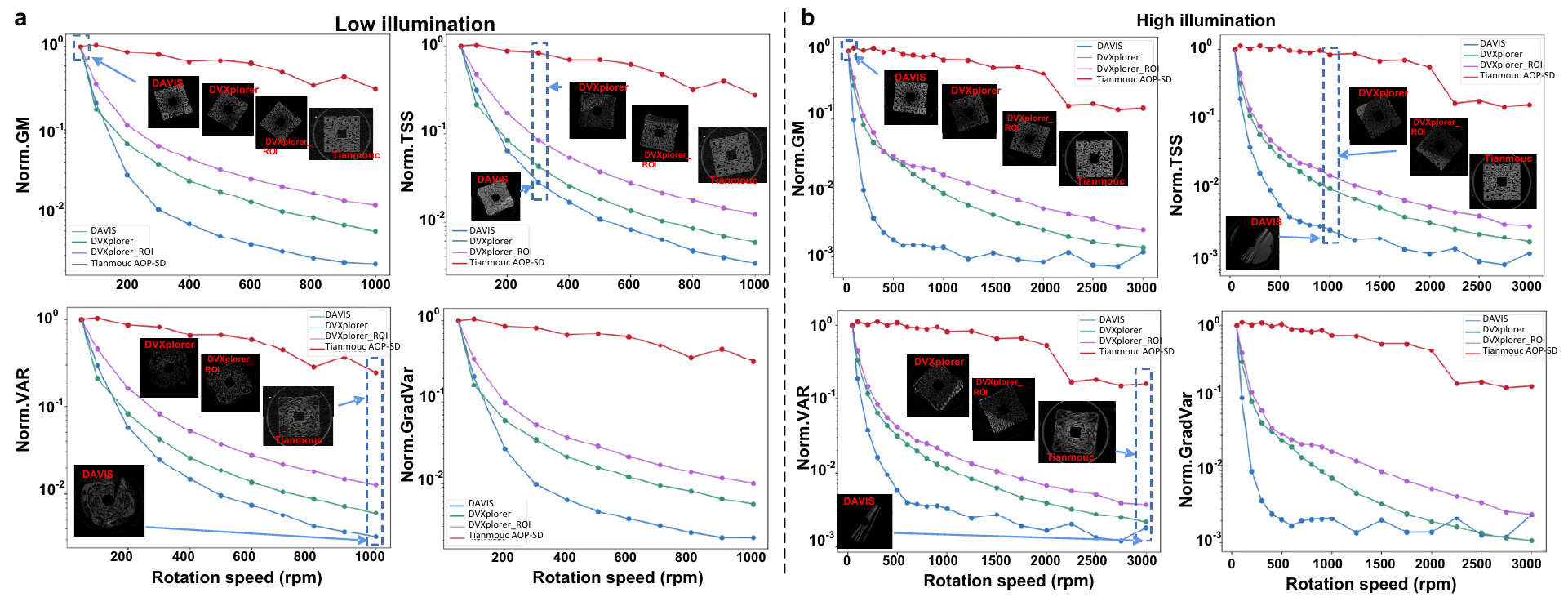} 
    \caption{Evaluation of structural information indicators under (a) low illumination and (b) high illumination.}
    \label{fig:edge}
\end{figure*}

As shown in Fig.\ref{fig:edge}, under low illumination conditions, the structural integrity of event data from DAVIS346, DVXplorer, and DVXplorer\_ROI deteriorates across all indicators as the object motion speed increases. In contrast, Tianmouc demonstrates strong robustness to speed fluctuations. Under high illumination, all three EVS continue to show a significant decline in structural indicators, whereas Tianmouc maintains high robustness. The degradation of structural information in EVS under low illumination is primarily attributed to increased pixel latency and reduced pixel bandwidth, leading to severe information loss when capturing patterns with complex features. Notably, DVXplorer\_ROI outperforms DAVIS and DVXplorer under most conditions, as its reduced event rate saturation mitigates the loss of structural information, though some degree of loss persists. This suggests that event rate saturation is the predominant factor causing distortion under high illumination conditions. In contrast, the global shutter operation of Tianmouc prevents event rate saturation, while the high-speed sampling mitigates temporal aliasing.

\subsection{Intensity reconstruction and frame interpolation }
To further investigate the impact of motion blur and structural information loss, we evaluate their effects on intensity reconstruction and frame interpolation, which are crucial for robotic perception because dataset labeling often relies on expert annotation on reconstructed frames \cite{he2024microsaccade}. For event-based intensity reconstruction, we adopt E2VID \cite{rebecq2019high}, which has demonstrated the highest no-reference quantitative performance across multiple scenarios, as identified by an event-based video reconstruction benchmark \cite{ercan2023evreal}. For Tianmouc AOP-SD data, we apply a classical Poisson blending algorithm \cite{tmc_code,sun2004poisson} to reconstruct grayscale intensity. The reconstructed video speed for all cameras is set to be the same. For frame interpolation, we adopt TimeLens \cite{tulyakov2021time} for RGB hybrid EVS and use the neural network-based frame interpolation algorithm reported in the original paper \cite{yang2024vision,tmc_code} for interpolating frames across all pathways in Tianmouc. The interpolation ratio of all cameras is set to be the same. Given the challenge of obtaining ground truth intensity values, we use the widely adopted no-reference metric, BRISQUE \cite{brisque}, as a quantitative indicator of intensity reconstruction quality.

\begin{figure}
    \centering
    \includegraphics[width=1\linewidth]{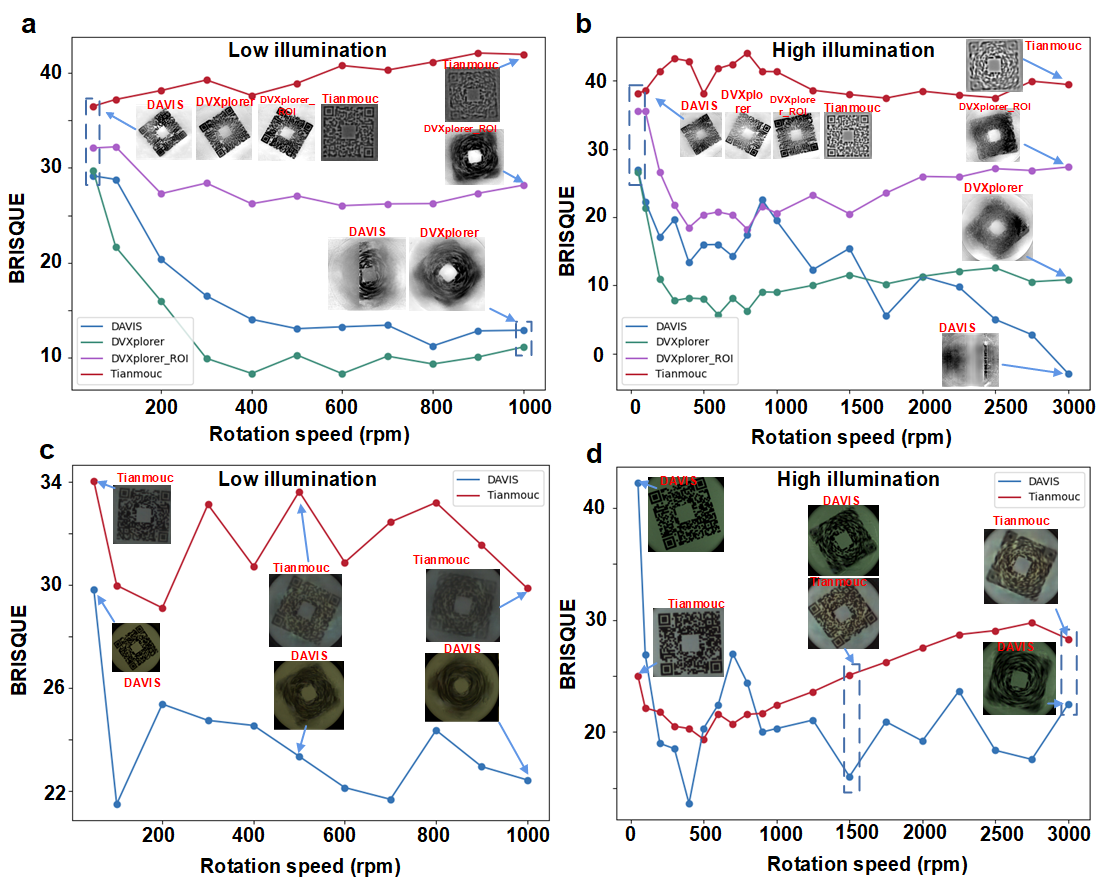} 
    \caption{ The evaluation of BRISQUE on grayscale intensity reconstruction from event or AOP-SD is shown in (a) for low illumination and (b) for high illumination. The evaluation of BRISUQE on colorful frame interpolation from RGB hybrid EVS in DAVIS346 or COP hybrid AOP in Tianmouc is shown in (c) for low illumination and (d) for high illumination.}
    \label{fig:recon}
\end{figure}

As illustrated in Fig.\ref{fig:recon}, the decline in BRISQUE scores aligns with the previously observed trends in structural information degradation. The intensity reconstruction quality of DVXplorer and DAVIS346, as well as the frame interpolation quality of DAVIS346, deteriorates as speed increases, whereas Tianmouc remains robust to speed variations.


\section{Robotic perception evaluation}
\label{sec:robo_perc}

In this section, we analyze the performance differences and trends in executing two important perception tasks, corner detection and motion estimation, across different cameras, from both spatial and temporal perspectives.
\subsection{Corner  detection}
\begin{figure}
    \centering
    \includegraphics[width=1\linewidth]{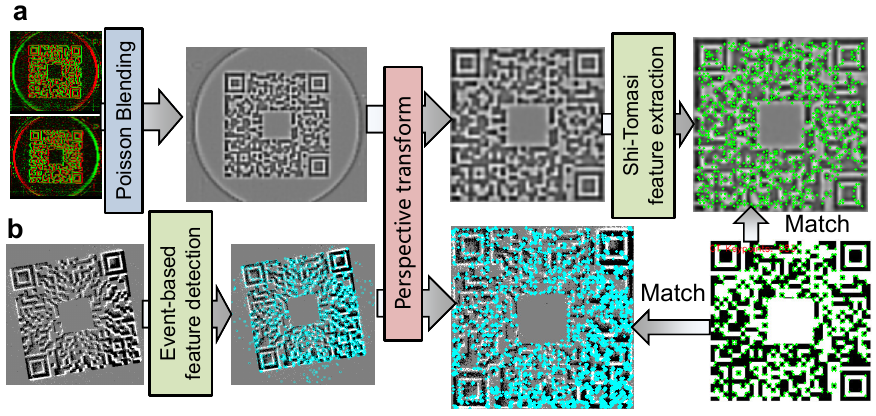} 
    \caption{ Corner detection evaluation procedure. (a) for Tianmouc and (b) for event-based corner detection.}
    \label{fig:feat_proc}
\end{figure}

Corner detection is fundamental in vision-based robotic systems, such as visual odometry. Fig.\ref{fig:feat_proc}a illustrates the corner detection process for Tianmouc. The AOP-SD data is first reconstructed into grayscale intensities using Poisson-blending, followed by a perspective transformation. This reconstruction step is highly practical for real-world applications, as the algorithm is computationally efficient, requiring approximately 1 ms on an embedded NVIDIA Jetson AGX Orin CPU. The transformed intensity data is then processed using Shi-Tomasi detection algorithms \cite{shi_tomasi}. Fig.\ref{fig:feat_proc}b presents the event-based corner detection pipeline. Raw event data collected over a time window corresponding to 1.5$^{\circ}$ of pattern rotation, which preserves clear edges, is first processed by ARC*\cite{arcstar}, a widely used event-based corner detection algorithm. The detected corners and raw data are then transformed using perspective transformation. However, due to the nature of ARC*, which generates multiple corner events for a single physical corner within a time window, significant overlap occurs in the transformed detection results. To mitigate this, we apply a 3-pixel neighborhood filter to remove redundant detections. Finally, the detected corners from Tianmouc’s AOP-SD and the transformed event-based detections are matched against ground truth corner positions, obtained using the Shi-Tomasi algorithms on original images.


\begin{table}[h]
\caption{Precision of corner detection match results}
\label{tab:prec}
\begin{tabular}{|p{0.4cm}|p{0.5cm}p{0.5cm}p{0.6cm}p{0.5cm}|p{0.5cm}p{0.5cm}p{0.6cm}p{0.5cm}|}
\hline
	 & \multicolumn{4}{c|}{ Low illumination}	&\multicolumn{4}{c|}{ High illumination}	\\
    \hline
RPM	& DAVIS& DVXp &	    DVXpl &	Tian&	 DAVIS&     DVXp &	    DVXpl &	Tian \\
    &      & lorer & orerROI & mouc&	 &      lorer & orerROI & mouc\\

\hline
50&	    0.454&	0.427&	0.516&	0.634&	0.472&	0.405&	0.485&	0.630\\
100	&   0.439&	0.440&	0.516&	0.638&	0.432&	0.473&	0.508&	0.638\\
200&	0.076&	0.192&	0.486&	0.627&	0.123&	0.220&	0.385&	0.639\\
300	&   N.A.&	0.191&	0.460&	0.628&	N.A.&	0.130&	0.187&	0.632\\
400	&   N.A.&	0.309&	0.379&	0.606&	N.A.&	0.180&	0.111&	0.633\\
500&	N.A.&	0.412&	0.417&	0.600&	N.A.&	0.199&	0.143&	0.636\\

\hline

\end{tabular}
\end{table}

\begin{table}[h]
\caption{Recall of corner detection match results}
\label{tab:recal}
\begin{tabular}{|p{0.4cm}|p{0.5cm}p{0.5cm}p{0.6cm}p{0.5cm}|p{0.5cm}p{0.5cm}p{0.6cm}p{0.5cm}|}
\hline
	 & \multicolumn{4}{c|}{ Low illumination}	&\multicolumn{4}{c|}{ High illumination}	\\
    \hline
RPM	& DAVIS& DVXp &	    DVXpl &	Tian&	 DAVIS&     DVXp &	    DVXpl &	Tian \\
    &      & lorer & orerROI & mouc&	 &      lorer & orerROI & mouc\\

\hline
50&	    0.667&	0.987&   0.909&	0.835& 0.742& 0.977& 0.899& 0.829\\
100&	0.686&	0.847&	0.886&	0.840& 0.682& 0.884& 0.914& 0.836\\
200&	0.072&  0.227&	0.826& 0.826& 0.085& 0.252& 0.537& 0.837\\
300&	N.A.&	0.147&	0.652& 0.827& N.A.& 0.107& 0.210& 0.833\\
400&	N.A.&	0.168&	0.433& 0.796& N.A.& 0.123& 0.099& 0.828\\
500&	N.A.&	0.125&	0.371& 0.781& N.A.& 0.105& 0.099& 0.829\\

\hline

\end{tabular}
\end{table}

\begin{table}[h]
\caption{F1 score of corner detection match results}\label{tab:f1}
\begin{tabular}{|p{0.4cm}|p{0.5cm}p{0.5cm}p{0.6cm}p{0.5cm}|p{0.5cm}p{0.5cm}p{0.6cm}p{0.5cm}|}
\hline
	 & \multicolumn{4}{c|}{ Low illumination}	&\multicolumn{4}{c|}{ High illumination}	\\
    \hline
RPM	& DAVIS& DVXp &	    DVXpl &	Tian&	 DAVIS&     DVXp &	    DVXpl &	Tian \\
    &      & lorer & orerROI & mouc&	 &      lorer & orerROI & mouc\\

\hline
50&    0.540&    0.596&    0.658&    0.720&    0.577&    0.573&    0.630&    0.716\\
100&    0.536&    0.579&    0.652&    0.725&    0.529&    0.616&    0.653&    0.724\\
200&    0.074&    0.208&    0.612&    0.713&    0.101&    0.235&    0.449&    0.725\\
300&    N.A.&    0.166&    0.540&    0.714&    N.A.&    0.117&    0.198&    0.719\\
400&    N.A.&    0.218&    0.404&    0.688&    N.A.&    0.146&    0.105&    0.717\\
500&    N.A.&    0.192&    0.393&    0.679&    N.A.&    0.137&    0.117&    0.720\\
\hline
\end{tabular}
\end{table}

The match $precision=N_{matched}/N_{Ndetect}$, $recall=N_{matched}/N_{GT}$ and F1 score results are summarized in Table.\ref{tab:prec}, \ref{tab:recal} and \ref{tab:f1} respectively. Consistent with the structural information degradation observed earlier, caused by low pixel bandwidth and event rate saturation, the number of detected and matched corners extracted from event data declines significantly as motion speed increases. In contrast, Tianmouc maintains high precision and robust performance across speed variations.


\subsection{Motion estimation}
\begin{figure}
    \centering
    \includegraphics[width=0.8\linewidth]{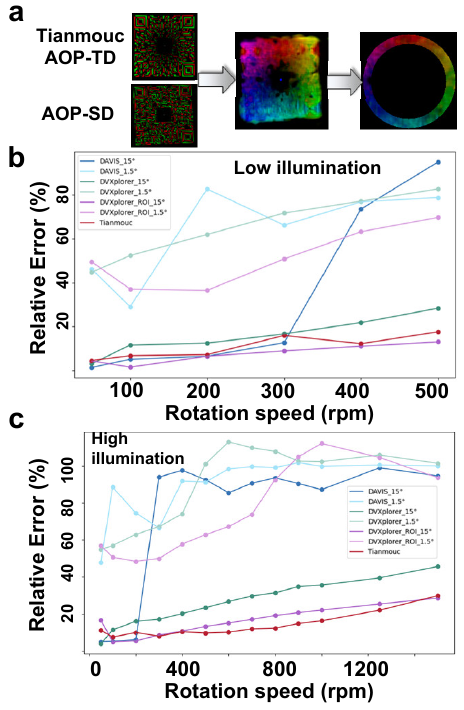} 
    \caption{ (a) The optical flow is masked by a ring for calculation of the relative error between estimated and GT angular speed. Relative error between estimated angular velocity and ground truth under (b) low illumination and (c) high illumination.}
    \label{fig:of}
\end{figure}

Motion estimation is a crucial temporal perception task in robotics. Here, we compare optical flow estimation using EVS and Tianmouc. Event-based optical flow is computed using a high-precision neural network-based algorithm \cite{zhu2018evflownet}, while optical flow from Tianmouc AOP-TD and AOP-SD is obtained using the approach described in \cite{yang2024vision}. For event-based asynchronous optical flow estimation, two temporal integration windows are employed, corresponding to 1.5$^{\circ}$ and 15$^{\circ}$ of pattern rotation. In contrast, Tianmouc utilizes a synchronized sampling scheme in which frame intervals are dynamically adjusted to align with rotational increments exceeding 1$^{\circ}$, ensuring consistent motion capture across speeds. To enable quantitative evaluation against ground truth, we extract angular velocity estimation from an annular region centered at 40\% to 50\% of the pattern’s total radius to avoid the high estimation error near the center of the pattern. This region minimizes radial distortion while preserving sufficient spatial resolution for accurate flow estimation. Using extended temporal windows, both sensors demonstrate robust motion estimation, with Tianmouc showing particular strength in preserving structural continuity and directional accuracy under high-speed conditions, as shown in Fig.\ref{fig:of}. These results highlight the accurate motion sensing capabilities of BVS technologies and their respective suitability for temporal tasks in robotic perception.

\section{Discussion and Conclusion}
\label{sec:discuss}
The preceding quantitative results provide a comprehensive comparison of different BVSs, highlighting their respective characteristics. All three evaluated BVSs exhibit competent performance in conventional scenarios and serve as effective complements to traditional cameras. EVS demonstrates high accuracy in scenarios with low feature complexity at high speeds or high feature complexity at moderate speeds, with the benefit of low bandwidth consumption. However, they encounter difficulty in accurate sensing at high speeds and highly complex scenes. The degradation of EVS imaging quality in high-speed motion may be attributed to several inherent limitation of EVS, including 1) pixel bandwidth variation under low-light conditions, in which limited photon collection reduces event generation and impairs perception; 2) event saturation, when rapid ego motion or object motion, especially in high-feature-complexity, overloads the on-chip bus and causes data loss; 3) data association and feature loss result from the purely temporal sensing scheme of EVS, in which object edges aligned with the camera’s motion direction may fail to generate detectable events, as previously reported\cite{he2024microsaccade}. While DAVIS346 can compensate for low-speed information loss by leveraging its RGB pathway, both RGB and event channels suffer degradation under high-speed motion. While employing larger temporal integration windows may alleviate issues in high-speed scenarios for EVS, similar to the approach used in motion estimation, it could also introduce higher latency. By contrast, Tianmouc’s AOP-SD exhibits a motion-robust, spatially gradient-sensitive data modality. Tianmouc’s combined encoding of complementary color intensity, TD, and SD, coupled with a high-speed sampling scheme, enables non-saturated, globally coherent acquisition of dynamic visual signals. This capability demonstrates its robustness in high-speed, cluttered environments, making it particularly suitable for robotic applications in such conditions. However, in low-light and ultra-high-speed scenarios, all sensors may experience motion blur due to the inherent limitations in pixel sensitivity, a fundamental constraint rooted in fabrication technology, which can improve as fabrication processes advance.

The field of BVS is still in a phase of rapid development and has yet to establish mature standards similar to those of traditional sensors. In our framework, we try our best to select representative and widely accepted post-processing algorithms for EVS, RGB hybrid EVS, and Tianmouc. Nevertheless, the development of event-based algorithms is advancing rapidly, with more improvements expanding EVS performance boundaries in robotic applications. For the system design, a more comprehensive test may require a thorough evaluation of more representative algorithms. Creating a comprehensive benchmark of cameras combined with algorithms will be a consideration for us in the future. As an emerging type of BVS technology, Tianmouc is highly promising for high-speed robotics. Nevertheless, its software and algorithms are still immature. Although integrating well-established traditional computer vision algorithms with Poisson-blending grayscale reconstruction of AOP-SD provides a viable approach to enhance usability, further algorithmic advancements are expected to process raw AOP data directly with low computational cost.


In our quantitative evaluations, we primarily focus on rotational motion, given its direct correlation with object motion speed on the focal plane (measured in pixels per second). Future research will expand this analysis to include complex movements, such as translational and combined movements.

In summary, we present a unified evaluation framework that bridges the modality gap among different BVS technologies, and quantitatively assess three representative BVSs, including an EVS, an RGB hybrid EVS, and Tianmouc. We experimentally demonstrate the strengths and weaknesses of these BVSs in terms of imaging quality and performance in core robotic vision tasks. Additionally, we elucidate the mechanisms underlying these limitations. These findings shed light on the application-dependent suitability of BVS technologies and support further advancements in this field. In future work, we will assess the performance of additional robotic tasks, such as visual odometry, and evaluate various BVS in high dynamic range environments.

    \bibliographystyle{IEEEtran}
   \bibliography{custom}

\begin{thebibliography}{10}
\providecommand{\url}[1]{#1}
\csname url@samestyle\endcsname
\providecommand{\newblock}{\relax}
\providecommand{\bibinfo}[2]{#2}
\providecommand{\BIBentrySTDinterwordspacing}{\spaceskip=0pt\relax}
\providecommand{\BIBentryALTinterwordstretchfactor}{4}
\providecommand{\BIBentryALTinterwordspacing}{\spaceskip=\fontdimen2\font plus
\BIBentryALTinterwordstretchfactor\fontdimen3\font minus \fontdimen4\font\relax}
\providecommand{\BIBforeignlanguage}[2]{{%
\expandafter\ifx\csname l@#1\endcsname\relax
\typeout{** WARNING: IEEEtran.bst: No hyphenation pattern has been}%
\typeout{** loaded for the language `#1'. Using the pattern for}%
\typeout{** the default language instead.}%
\else
\language=\csname l@#1\endcsname
\fi
#2}}
\providecommand{\BIBdecl}{\relax}
\BIBdecl

\bibitem{gallego2020event}
G.~Gallego, T.~Delbr{\"u}ck, G.~Orchard, C.~Bartolozzi, B.~Taba, A.~Censi, S.~Leutenegger, A.~J. Davison, J.~Conradt, K.~Daniilidis \emph{et~al.}, ``Event-based vision: A survey,'' \emph{IEEE transactions on pattern analysis and machine intelligence}, vol.~44, no.~1, pp. 154--180, 2020.

\bibitem{yang2024vision}
Z.~Yang, T.~Wang, Y.~Lin, Y.~Chen, H.~Zeng, J.~Pei, J.~Wang, X.~Liu, Y.~Zhou, J.~Zhang \emph{et~al.}, ``A vision chip with complementary pathways for open-world sensing,'' \emph{Nature}, vol. 629, no. 8014, pp. 1027--1033, 2024.

\bibitem{evk4spec}
Prophesee, ``Evk4 hd camera manual,'' https://www.prophesee.ai/wp-content/uploads/2023/01/EVK4-HD-Prophesee-Evaluation-Kit-Camera-Manual-1.1.pdf.

\bibitem{inivation_all_spec}
iniVation, ``Specifications – current models,'' https://inivation.com/wp-content/uploads/2021/08/2021-08-iniVation-devices-Specifications.pdf.

\bibitem{brandli2014240}
C.~Brandli, R.~Berner, M.~Yang, S.-C. Liu, and T.~Delbruck, ``A 240$\times$ 180 130 db 3 $\mu$s latency global shutter spatiotemporal vision sensor,'' \emph{IEEE Journal of Solid-State Circuits}, vol.~49, no.~10, pp. 2333--2341, 2014.

\bibitem{DAVIS346_spec}
iniVation, ``Specifications – davis 346,'' https://inivation.com/wp-content/uploads/2023/03/DAVIS346.pdf.

\bibitem{he2024microsaccade}
B.~He, Z.~Wang, Y.~Zhou, J.~Chen, C.~D. Singh, H.~Li, Y.~Gao, S.~Shen, K.~Wang, Y.~Cao \emph{et~al.}, ``Microsaccade-inspired event camera for robotics,'' \emph{Science Robotics}, vol.~9, no.~90, p. eadj8124, 2024.

\bibitem{jahne2010emva}
B.~J{\"a}hne, ``Emva 1288 standard for machine vision: Objective specification of vital camera data,'' \emph{Optik \& Photonik}, vol.~5, no.~1, pp. 53--54, 2010.

\bibitem{tapia2023comparison}
R.~Tapia, J.~P. Rodr{\'\i}guez-G{\'o}mez, J.~A. Sanchez-Diaz, F.~J. Ga{\~n}{\'a}n, I.~G. Rodr{\'\i}guez, J.~Luna-Santamaria, J.~Mart{\'\i}nez-De~Dios, and A.~Ollero, ``A comparison between framed-based and event-based cameras for flapping-wing robot perception,'' in \emph{2023 IEEE/RSJ International Conference on Intelligent Robots and Systems (IROS)}.\hskip 1em plus 0.5em minus 0.4em\relax IEEE, 2023, pp. 3025--3032.

\bibitem{holevsovsky2021experimental}
O.~Hole{\v{s}}ovsk{\`y}, R.~{\v{S}}koviera, V.~Hlav{\'a}{\v{c}}, and R.~Vitek, ``Experimental comparison between event and global shutter cameras,'' \emph{Sensors}, vol.~21, no.~4, p. 1137, 2021.

\bibitem{hu2021v2e}
Y.~Hu, S.-C. Liu, and T.~Delbruck, ``v2e: From video frames to realistic dvs events,'' in \emph{Proceedings of the IEEE/CVF conference on computer vision and pattern recognition}, 2021, pp. 1312--1321.

\bibitem{gehrig2022high}
D.~Gehrig and D.~Scaramuzza, ``Are high-resolution event cameras really needed?'' \emph{arXiv preprint arXiv:2203.14672}, 2022.

\bibitem{yang2024latency}
Y.~Yang, J.~Liang, B.~Yu, Y.~Chen, J.~S. Ren, and B.~Shi, ``Latency correction for event-guided deblurring and frame interpolation,'' in \emph{Proceedings of the IEEE/CVF Conference on Computer Vision and Pattern Recognition}, 2024, pp. 24\,977--24\,986.

\bibitem{low2024deblur}
W.~F. Low and G.~H. Lee, ``Deblur e-nerf: Nerf from motion-blurred events under high-speed or low-light conditions,'' in \emph{European Conference on Computer Vision}.\hskip 1em plus 0.5em minus 0.4em\relax Springer, 2024, pp. 192--209.

\bibitem{gehrig2020eklt}
D.~Gehrig, H.~Rebecq, G.~Gallego, and D.~Scaramuzza, ``Eklt: Asynchronous photometric feature tracking using events and frames,'' \emph{International Journal of Computer Vision}, vol. 128, no.~3, pp. 601--618, 2020.

\bibitem{ercan2023evreal}
B.~Ercan, O.~Eker, A.~Erdem, and E.~Erdem, ``Evreal: Towards a comprehensive benchmark and analysis suite for event-based video reconstruction,'' in \emph{Proceedings of the IEEE/CVF Conference on Computer Vision and Pattern Recognition}, 2023, pp. 3942--3951.

\bibitem{ding2023mlb}
S.~Ding, J.~Chen, Y.~Wang, Y.~Kang, W.~Song, J.~Cheng, and Y.~Cao, ``E-mlb: Multilevel benchmark for event-based camera denoising,'' \emph{IEEE Transactions on Multimedia}, 2023.

\bibitem{duan2023eventaid}
P.~Duan, B.~Li, Y.~Yang, H.~Lou, M.~Teng, Y.~Ma, and B.~Shi, ``Eventaid: Benchmarking event-aided image/video enhancement algorithms with real-captured hybrid dataset,'' \emph{arXiv preprint arXiv:2312.08220}, 2023.

\bibitem{gao2023action}
Y.~Gao, J.~Lu, S.~Li, N.~Ma, S.~Du, Y.~Li, and Q.~Dai, ``Action recognition and benchmark using event cameras,'' \emph{IEEE Transactions on Pattern Analysis and Machine Intelligence}, 2023.

\bibitem{wang2024event}
X.~Wang, S.~Wang, C.~Tang, L.~Zhu, B.~Jiang, Y.~Tian, and J.~Tang, ``Event stream-based visual object tracking: A high-resolution benchmark dataset and a novel baseline,'' in \emph{Proceedings of the IEEE/CVF Conference on Computer Vision and Pattern Recognition}, 2024, pp. 19\,248--19\,257.

\bibitem{movementvsDVS}
J.~Barrios-Avil{\'e}s, T.~Iakymchuk, J.~Samaniego, L.~D. Medus, and A.~Rosado-Mu{\~n}oz, ``Movement detection with event-based cameras: comparison with frame-based cameras in robot object tracking using powerlink communication,'' \emph{Electronics}, vol.~7, no.~11, p. 304, 2018.

\bibitem{censi2015power}
A.~Censi, E.~Mueller, E.~Frazzoli, and S.~Soatto, ``A power-performance approach to comparing sensor families, with application to comparing neuromorphic to traditional vision sensors,'' in \emph{2015 IEEE International Conference on Robotics and Automation (ICRA)}.\hskip 1em plus 0.5em minus 0.4em\relax IEEE, 2015, pp. 3319--3326.

\bibitem{iwe2018}
G.~Gallego, H.~Rebecq, and D.~Scaramuzza, ``A unifying contrast maximization framework for event cameras, with applications to motion, depth, and optical flow estimation,'' in \emph{Proceedings of the IEEE conference on computer vision and pattern recognition}, 2018, pp. 3867--3876.

\bibitem{eros}
G.~Goyal, F.~Di~Pietro, N.~Carissimi, A.~Glover, and C.~Bartolozzi, ``Moveenet: online high-frequency human pose estimation with an event camera,'' in \emph{Proceedings of the IEEE/CVF Conference on Computer Vision and Pattern Recognition}, 2023, pp. 4024--4033.

\bibitem{yang2016analysis}
M.~Yang, S.-C. Liu, and T.~Delbruck, ``Analysis of encoding degradation in spiking sensors due to spike delay variation,'' \emph{IEEE Transactions on Circuits and Systems I: Regular Papers}, vol.~64, no.~1, pp. 145--155, 2016.

\bibitem{rebecq2019high}
H.~Rebecq, R.~Ranftl, V.~Koltun, and D.~Scaramuzza, ``High speed and high dynamic range video with an event camera,'' \emph{IEEE transactions on pattern analysis and machine intelligence}, vol.~43, no.~6, pp. 1964--1980, 2019.

\bibitem{tmc_code}
\BIBentryALTinterwordspacing
T.~Wang, ``Code of "a vision chip with complementary pathways for open-world sensing",'' Mar. 2024. [Online]. Available: \url{https://doi.org/10.5281/zenodo.10775253}
\BIBentrySTDinterwordspacing

\bibitem{sun2004poisson}
J.~Sun, J.~Jia, C.-K. Tang, and H.-Y. Shum, ``Poisson matting,'' \emph{ACM Transactions on Graphics (TOG)}, vol.~23, no.~3, pp. 315--321, 2004.

\bibitem{tulyakov2021time}
S.~Tulyakov, D.~Gehrig, S.~Georgoulis, J.~Erbach, M.~Gehrig, Y.~Li, and D.~Scaramuzza, ``Time lens: Event-based video frame interpolation,'' in \emph{Proceedings of the IEEE/CVF conference on computer vision and pattern recognition}, 2021, pp. 16\,155--16\,164.

\bibitem{brisque}
A.~Mittal, A.~K. Moorthy, and A.~C. Bovik, ``No-reference image quality assessment in the spatial domain,'' \emph{IEEE Transactions on image processing}, vol.~21, no.~12, pp. 4695--4708, 2012.

\bibitem{shi_tomasi}
J.~Shi \emph{et~al.}, ``Good features to track,'' in \emph{1994 Proceedings of IEEE conference on computer vision and pattern recognition}.\hskip 1em plus 0.5em minus 0.4em\relax IEEE, 1994, pp. 593--600.

\bibitem{arcstar}
I.~Alzugaray and M.~Chli, ``Asynchronous corner detection and tracking for event cameras in real time,'' \emph{IEEE Robotics and Automation Letters}, vol.~3, no.~4, pp. 3177--3184, 2018.

\bibitem{zhu2018evflownet}
A.~Z. Zhu and L.~Yuan, ``Ev-flownet: Self-supervised optical flow estimation for event-based cameras,'' in \emph{Robotics: Science and Systems}, 2018.

\end{thebibliography}

\end{document}